\documentclass{bmvc2k}

\usepackage{amsmath,amsfonts,amssymb}
\usepackage{graphicx}
\usepackage{booktabs}   
\usepackage{multirow}
\usepackage{siunitx}    
\usepackage{url}
\usepackage{float}
\usepackage{tabularx}
\usepackage[table]{xcolor}
\newcolumntype{C}{>{\centering\arraybackslash}X}
\usepackage{calc}
\usepackage{caption}       
\usepackage{subcaption}    
\usepackage{array}
\hypersetup{hidelinks}

\newcommand{\datasetName}{\textsc{OHL-UK}}
\newcommand{\doi}[1]{\href{https://doi.org/#1}{\nolinkurl{#1}}}

\newcommand{\model}[1]{\textsc{#1}}
\newcommand{\yoloviii}{\model{YOLOv8}}
\newcommand{\dinodetr}{\model{DINO-DETR}}
\newcommand{\detr}{\model{DETR}}


\usepackage{placeins} 
\title{Advancing Utility Pole and Sign
Detection Through Deep Learning}

\addauthor{Carl Dickinson}{carl.dickinson@strath.ac.uk}{1}
\addauthor{Gaetano Di Caterina}{gaetano.di-caterina@strath.ac.uk}{1}

\addinstitution{
University of Strathclyde\\
United Kingdom
}

\runninghead{Dickinson \& Di Caterina}{Utility Pole and Warning Sign Detection}

\begin{document}
\maketitle

\begin{abstract}
Utility poles are an essential part of the infrastructure used to support power distribution systems and other critical public services. Their regular inspection is crucial to ensure the stability and safety of the electrical grid.
A deep learning framework is presented for the automated detection, segmentation and lean angle estimation of wooden utility poles, and classification of attached electrical warning signs, using ground-level imagery. The system is trained on a custom dataset of 4,570 annotated images extracted from Google Street View, featuring challenging real-world scenes with visually ambiguous wooden poles lacking distinctive features. The proposed model is based on the Detection Transformer (DETR), suitably modified and trained on the custom dataset. The model outperforms standard object detectors (RetinaNet, Faster R-CNN, YOLOv3-Tiny), achieving a mean average precision of 90.43\% for pole detection and 88.26\% for sign detection.
Extending this model with a segmentation head enables per-instance mask generation, which is then used to estimate pole lean angle. The model accurately estimates lean for 1,367 out of 1,433 test-set poles, with a mean absolute error of 1.01$^\circ$. Moreover, the custom dataset created in this work is also made publicly available to be used as a benchmark.
\end{abstract}

\section{Introduction}
\label{sec:introduction}
A reliable electricity supply hinges on the condition of overhead-line (OHL) infrastructure.  
In Great Britain alone, wooden poles carry more than \SI{800000}{km} of distribution cables \cite{UKPN2017-FutureSmart}.  
To survey this infrastructure, annual foot patrols and costly helicopter surveys remain the norm, yet they are hazardous, slow, and subjective.  
Attempts to automate inspection with helicopters \cite{Jones2001}, UAVs \cite{Sharma2015}, or climbing robots \cite{Goncalves2013} have been hindered by camera stabilisation, obstacle traversal, and limited computer-vision accuracy.  
Recent object detectors based on Convolutional Neural Networks (CNN) (e.g.\ RetinaNet, Faster R-CNN, YOLO) improve pole localisation \cite{Liu2019,UsingDL}, but struggle with visually ambiguous wooden poles without crossarms and provide no direct geometric descriptors such as lean angle.

This paper presents a unified detection–segmentation pipeline, based on the Detection Transformer (DETR) \cite{Carion2020}, that both \emph{detects} wooden poles and warning signs and \emph{estimates} pole lean angle from a single Google Street View (GSV) image.  
To support training and evaluation, the first open dataset of its kind has also been curated in this work: 4,570 GSV images containing 6,773 poles and 1,805 signs, partitioned into \num{2920}/\num{730}/\num{920} train/val/test images. 
We release \datasetName{} with trained checkpoints \cite{ohluk2025}.

The fine-tuned DETR model achieves a mean average precision (mAP) of \SI{90.43}{\percent} for pole detection and \SI{88.26}{\percent} for sign detection, surpassing RetinaNet, YOLOv3-Tiny, and Faster R-CNN baselines.  
By attaching a lightweight segmentation head, the model also estimates pole lean angle with a mean absolute error of \SI{1.01}{\degree}; \SI{98}{\percent} of predictions lie within \SI{5}{\degree}, and \SI{90}{\percent} within \SI{2}{\degree}, outperforming prior work by up to \SI{22.3}{\percent}.

The contribution of this work is three-fold:
\begin{enumerate}
  \item \textbf{OHL-UK Dataset}: the first publicly shareable GSV corpus of wooden poles \emph{without} crossarms, with bounding-box and pixel-level annotations.

  \item \textbf{DETR-based detection model}: state-of-the-art accuracy on pole and warning-sign detection.
  \item \textbf{Segmentation extension for lean-angle estimation}: integrated in the same network, delivering \SI{1.01}{\degree} MAE without multi-stage heuristics.
\end{enumerate}

The remainder of the paper is organised as follows. Section~\ref{sec:method} details the dataset creation, augmentation pipeline, model architecture, and lean-angle estimation strategy.  
Section~\ref{sec:experiments} presents a comprehensive empirical evaluation, including detection benchmarking, segmentation accuracy, and ablation studies.  
Section~\ref{sec:discussion} outlines the system’s limitations and technical challenges observed during experimentation.  
Section~\ref{sec:conclusion} summarises key contributions and suggests directions for future research.

\section{Methodology}
\label{sec:method}
\subsection{OHL-UK Dataset: Wooden Utility Pole and Sign Corpus}
\label{sec:dataset}
To support the development and evaluation of the detection and lean estimation framework, a large-scale dataset of wooden utility poles and attached electrical warning signs was curated. The dataset, termed \textbf{OHL-UK}, was constructed using ground-level imagery obtained via the Google Street View (GSV) API \cite{GSVAPI}. Collection was guided by over 670,000 geographic coordinates provided by UK Power Networks.
For each coordinate, four images were captured at compass headings of 0$^\circ$, 90$^\circ$, 180$^\circ$, and 270$^\circ$ to maximise pole visibility. Images were cropped to 640$\times$640 pixels and manually filtered to remove irrelevant scenes.

Two object classes were annotated, namely
    \textbf{Wooden utility poles}, including visually ambiguous poles without crossarms;
    and \textbf{Electrical warning signs}, typically affixed to poles and occupying small pixel regions.

Annotations were generated using the VIA tool \cite{Dutta2019} and stored in COCO-style JSON format, including both bounding boxes and polygonal segmentations. Each pole annotation is additionally labelled with a lean angle in degrees, computed by fitting a line to the segmentation mask using OpenCV's \texttt{fitLine()} method.

To improve generalisation during training, a set of data augmentations was applied, including horizontal flips, random resizes, size-crops, colour jitter, histogram equalisation, and rotations. These augmentations follow schemes proposed in \cite{Cubuk2019, Mounsaveng2021}, adapted for ImageNet-pretrained backbones.

\begin{figure}[t]
\centering
\begin{tabular}{cc}
\bmvaHangBox{\includegraphics[height=4cm]{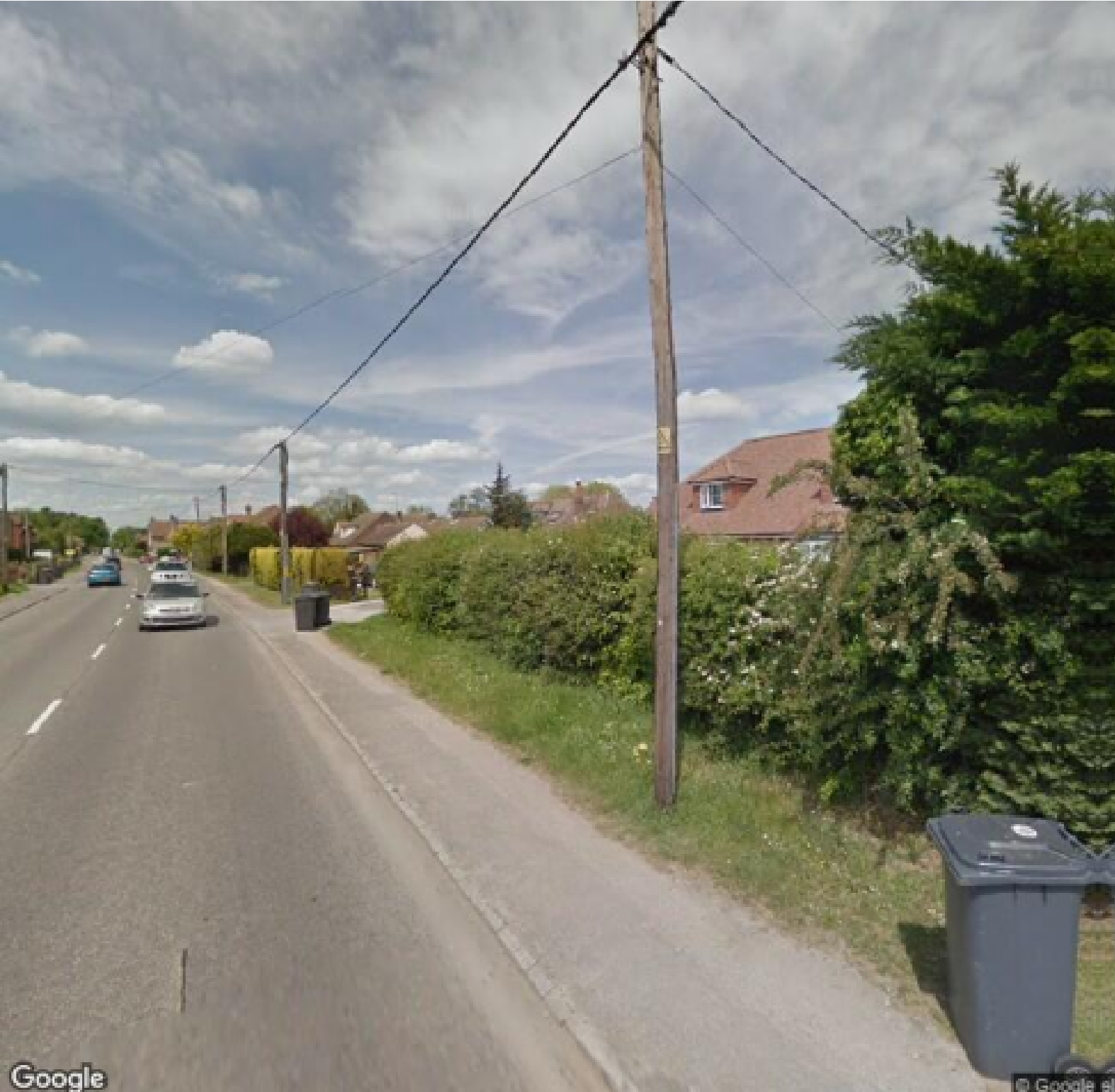}} &
\bmvaHangBox{\includegraphics[height=4cm]{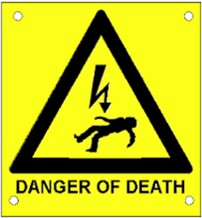}} \\
(a) Wooden utility pole & (b) Danger of Death sign
\end{tabular}
\vspace{0.6em}
\caption{Examples of a wooden utility pole and an electrical warning sign from the OHL-UK dataset.}
\label{fig:utility_and_sign}
\end{figure}

A breakdown of dataset composition, annotation types, and class distributions is summarised in Table~\ref{tab:dataset_stats}. Figure~\ref{fig:utility_and_sign} illustrates representative examples of the two object classes present in the dataset.

\begin{table}[t]
\centering
\caption{Summary statistics of the OHL-UK dataset.}
\vspace{0.5em}
\label{tab:dataset_stats}
\begin{tabular}{ll}
\toprule
Total images & 4,570 \\
Image resolution & 640$\times$640 pixels \\
Object classes & Wooden utility pole, Electrical warning sign \\
Annotation types & Bounding box, polygonal mask, lean angle \\
Annotated poles & 6,773 \\
Annotated signs & 1,805 \\
Train/Validation/Test split & 2,920 / 730 / 920 images \\
\hspace{1em} \textbullet~Poles per split & 4,269 / 1,068 / 1,490 \\
\hspace{1em} \textbullet~Signs per split & 1,008 / 250 / 552 \\
Data source & Google Street View API \\
Annotation format & COCO JSON (via VIA tool) \\
Augmentation types & Flip, resize, crop, jitter, equalisation, rotation \\
Dataset availability & Public release upon paper acceptance \\
\bottomrule
\end{tabular}
\end{table}

\subsection{Model Architecture}
\label{sec:model_architecture}
\begin{figure}[t]
\centering
\includegraphics[width = 8cm]{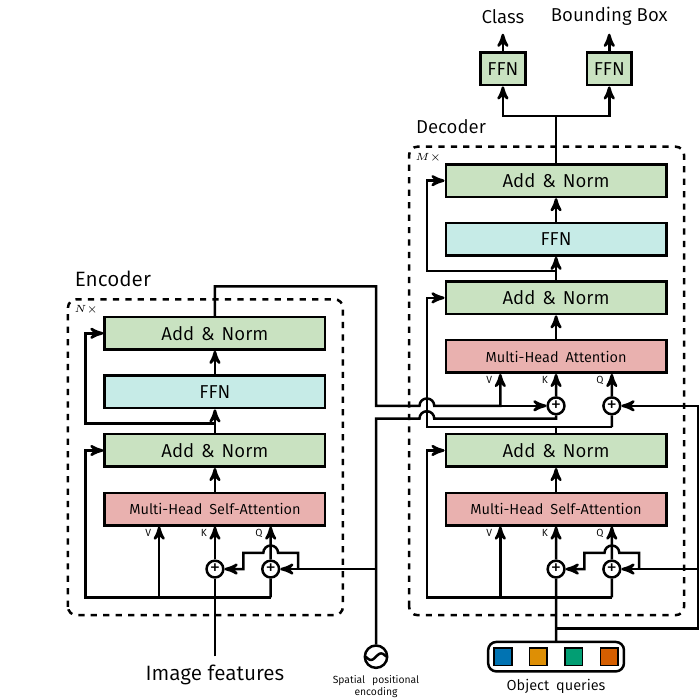}
\vspace{0.6em}
\caption{DETR architecture: a convolutional backbone extracts features from the input image, which are processed by a transformer encoder–decoder to generate object predictions. Adapted from \cite{Carion2020}.}
\label{fig:detr_architecture}
\end{figure}

The object detection pipeline is based on the DEtection TRansformer (DETR) \cite{Carion2020} (Figure~\ref{fig:detr_architecture}), which combines a ResNet-50 convolutional backbone with a transformer encoder–decoder and a bipartite matching loss.
 Unlike traditional detectors, DETR eliminates the need for hand-crafted anchors and non-maximum suppression by directly predicting a fixed set of object queries.
The model is adapted to a binary classification task for detecting wooden poles and warning signs. It is fine-tuned using the AdamW optimiser with focal loss terms to address class imbalance.

\begin{figure}[t]
\centering
\includegraphics[width=\linewidth]{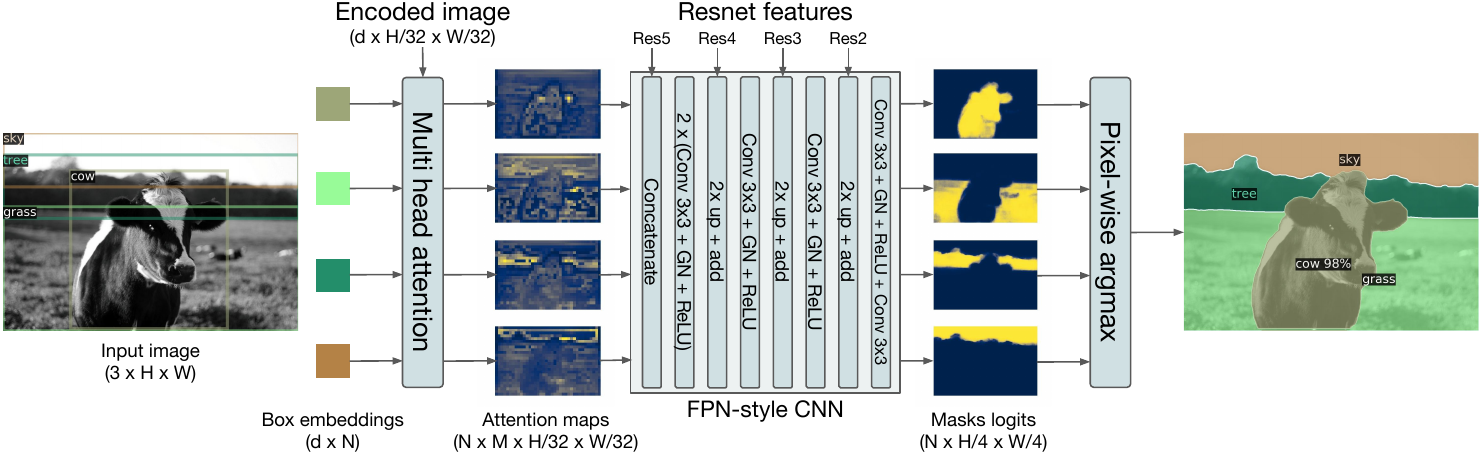}
\vspace{0.6em}
\caption{Segmentation head added to DETR, enabling binary mask prediction for each detected utility pole. Adapted from \cite{Carion2020}.}
\label{fig:panoptic_head}
\end{figure}

\label{sec:segmentation}
To support per-instance orientation estimation, the base DETR model is extended with a segmentation head. This module enables the generation of dense binary masks for each detected object, including visually ambiguous wooden utility poles. Figure~\ref{fig:panoptic_head} shows the overall structure of the segmentation extension.

The segmentation head attaches to the output of the transformer decoder and produces a high-resolution mask for each object. This design follows the panoptic extension described in \cite{Carion2020}, incorporating multi-head attention and feature upsampling to enable accurate instance-wise masks.

Estimated pole masks are post-processed using OpenCV’s \texttt{fitLine()} function to obtain the dominant orientation vector. From this, the lean angle \( \theta \) is computed as:
\begin{equation}
\theta = \arctan\left(\frac{\text{rise}}{\text{run}}\right)
\end{equation}
Here, \emph{rise} and \emph{run} refer to the vertical and horizontal components, respectively, of the line of best fit through the segmented mask. Each angle is recorded in degrees and appended to the corresponding pole annotation as supplementary metadata. The full model is trained end-to-end, with classification, box, and mask losses optimised jointly.

\subsection{Hardware and Training Setup}
Training and evaluation were performed on the ARCHIE-WeSt HPC \cite{Archie_West} at the University of Strathclyde using an NVIDIA A100 40GB GPU. The software stack included Python 3.9, PyTorch, TorchVision, and Keras. DETR was initialised with ImageNet-pretrained weights. All experiments were scripted for reproducibility.

\section{Baseline Experiments}
\label{sec:experiments}
This section presents a comprehensive evaluation of the object detection and segmentation components of our framework. We benchmark the DETR model against established deep learning detectors—RetinaNet, YOLO v3-Tiny, and Faster R-CNN—using consistent hyperparameters, dataset partitions, and evaluation metrics. Additionally, we assess the extended DETR + segmentation architecture for its ability to estimate pole lean angle with high precision.

\subsection{Common Experimental Setup}
All models were trained on a dataset of 2,920 images, validated on 730 images, and evaluated on a held-out test set of 920 images containing 1,490 utility poles and 552 warning signs. Each model underwent a 5-fold cross-validation, using stratified splits to maintain class distribution. Evaluation metrics include mAP, true/false positives and negatives, F1 scores, and  Intersection-over-Union (IoU) detection thresholds.

The following hyperparameters were applied uniformly unless model-specific constraints required modification:
\begin{itemize}
    \item Batch size: 1 (RetinaNet, YOLO v3-Tiny, Faster R-CNN); 4 (DETR)
    \item Epoch count: 200 (RetinaNet, YOLO v3-Tiny, Faster R-CNN); 300 (DETR)
    \item Optimiser: Adam (RetinaNet, YOLO v3-Tiny); SGD (Faster R-CNN); AdamW (DETR)
    \item Augmentations: horizontal flip, resizing, cropping, rotation, colour jitter, equalisation
    \item Cross-validation folds: 5
\end{itemize}

To optimise detection performance, an ablation study was performed on the DETR model by varying the learning rate, weight decay, and data augmentation combinations. Each configuration was trained for 300 epochs, and mAP was recorded for poles and signs. The best result---mean mAP of 90.46\%---was achieved with a learning rate of \(1 \times 10^{-5}\), weight decay of \(1 \times 10^{-4}\), and the following data augmentations: random horizontal flip, colour jitter, equalisation, random resize, random size crop, and rotation. This configuration was used for all subsequent experiments.

Each model’s optimal configuration was identified via cross-validation. The best checkpoint was then evaluated on the held-out test set. Performance was reported at IoU thresholds from 0.0 to 1.0, capturing detection robustness across a range of spatial precision requirements.

\subsection{Cross-Validation Detection Performance}
All four object detection models were benchmarked using five-fold cross-validation. Table~\ref{tab:model_comparison} reports mAP achieved for pole detection, along with the epoch at which each model peaked.

\begin{table}[t]
\centering
\footnotesize
\begin{tabular}{lcc}
\toprule
\textbf{Model} & \textbf{Pole mAP} & \textbf{Peak Epoch} \\
\midrule
DETR & 90\% & 91 \\
YOLO v3-Tiny & 77\% & 12 \\
Faster R-CNN & 64\% & 24 \\
RetinaNet & 23\% & 9 \\
\bottomrule
\addlinespace[1.9ex]
\end{tabular}
\caption{Cross-validation performance on wooden utility pole detection (mean mAP over five folds).}
\label{tab:model_comparison}
\end{table}

RetinaNet performed the weakest, with poor convergence and low precision across folds. YOLO v3-Tiny trained rapidly and consistently, achieving moderate accuracy with early stopping. Faster R-CNN benefited from its two-stage architecture, but showed higher variance across splits. DETR outperformed all baselines, reaching 90\% mAP with low variance and strong generalisation. Its transformer-based attention enabled robust localisation despite challenging pole appearances and cluttered backgrounds.
These results informed the final configuration selection for test set evaluation.

\begin{figure}[t]
\centering
\includegraphics[width=0.4\linewidth]{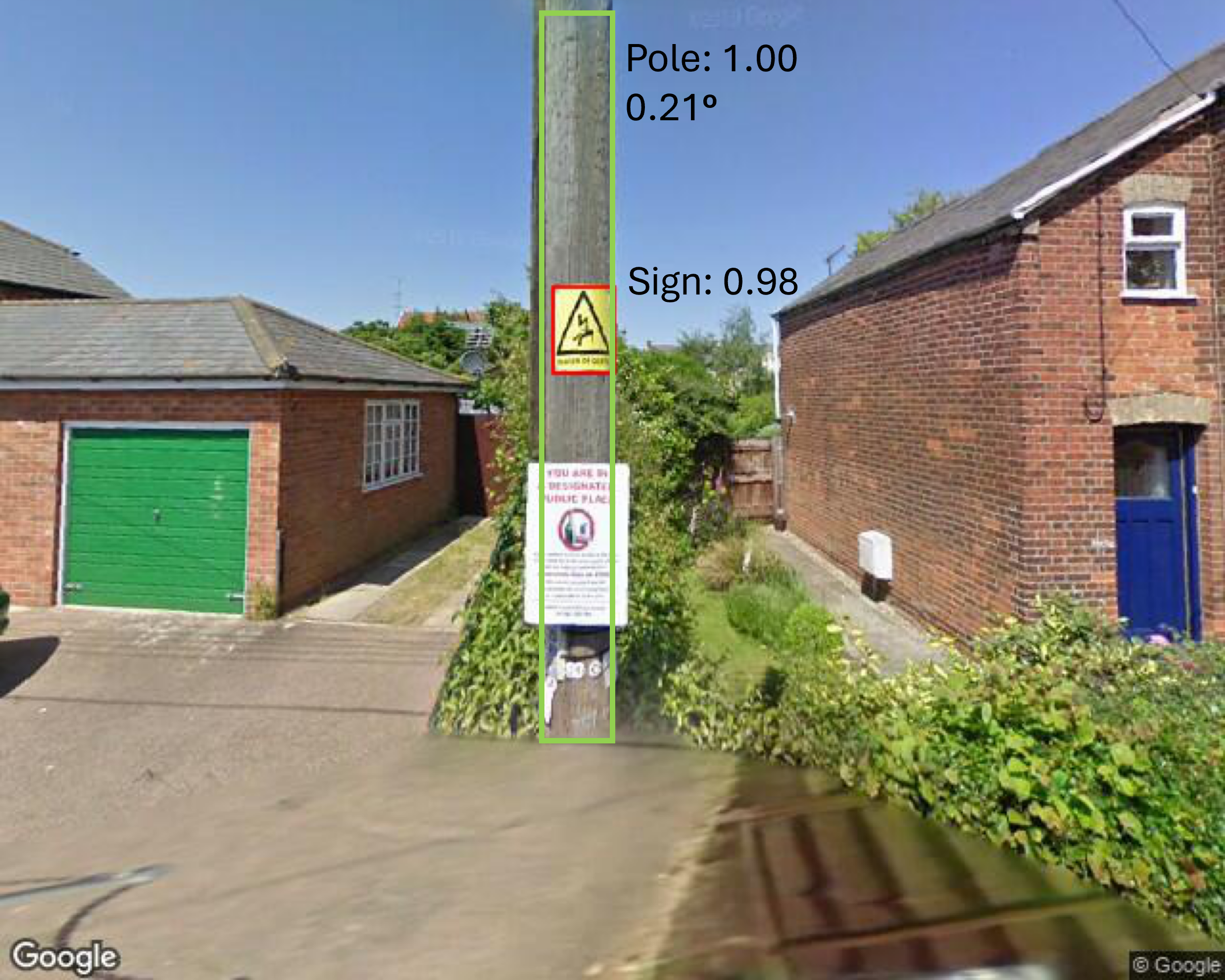}
\vspace{0.6em}
\caption{Example DETR detection output showing confident localisation of a wooden utility pole and associated warning sign.}
\label{fig:pole_sign_detection}
\end{figure}

An example detection output is shown in Figure~\ref{fig:pole_sign_detection}, where DETR correctly identifies both a wooden utility pole and an attached warning sign, with high confidence scores.

\subsection{Test Set Analysis of Detection Outcomes and Performance Trends}
Table~\ref{tab:Test-Comparison} presents detection results on the held-out test set, comprising 920 images with 6,773 wooden utility poles and 1,805 warning signs. The table reports true positives (TP), false positives (FP), false negatives (FN), and F1 scores across IoU thresholds from 0.0 to 1.0 for each model.

Across all thresholds, \textbf{DETR records the highest number of true positives}, beginning with 1,433 at an IoU of 0.0 and maintaining superiority as spatial constraints increase. Faster R-CNN and YOLO follow, while RetinaNet consistently underperforms—highlighting challenges in precise pole localisation.

\textbf{False positive counts rise} with increasing IoU thresholds for all models. DETR begins with only 86 FPs at IoU~0.0, escalating to 1,495 at IoU~1.0. YOLO and Faster R-CNN follow similar trajectories but exhibit higher baseline rates.

\textbf{False negatives remain lowest for DETR}, indicating strong recall, particularly in lenient evaluation regimes. RetinaNet again fares the worst, producing the highest FN rates throughout.

In terms of F1 performance, \textbf{DETR peaks at 0.95 at IoU~0.0} and sustains leading performance as thresholds rise, although all models degrade under stricter criteria. YOLO and Faster R-CNN remain competitive; RetinaNet consistently lags behind.

Overall, \textbf{DETR demonstrates strong test set robustness}, especially under low- to mid-stringency IoU thresholds. Like other models, however, its localisation precision deteriorates under stricter conditions, revealing inherent trade-offs in transformer-based global reasoning.

    \begin{table}[t]
    \centering
    \setlength{\tabcolsep}{3.5pt}
    \footnotesize
    \begin{tabularx}{\textwidth}{p{1.3cm}|C|C|C|C|C|C|C|C|C|C|C|C}
    \toprule
    \textbf{Model} & \textbf{IoU} & \textbf{0.0} & \textbf{0.1} & \textbf{0.2} & \textbf{0.3} & \textbf{0.4} & \textbf{0.5} & \textbf{0.6} & \textbf{0.7} & \textbf{0.8} & \textbf{0.9} & \textbf{1.0} \\
    \midrule
    \textbf{DETR} & \multirow{4}{*}{\textbf{TP}} & \cellcolor{yellow}1,433 & \cellcolor{yellow}1,431 & \cellcolor{yellow}1,428 & \cellcolor{yellow}1,398 & \cellcolor{yellow}1,344 & \cellcolor{yellow}1,295 & \cellcolor{yellow}1,214 & \cellcolor{yellow}1,041 & \cellcolor{yellow}746 & \cellcolor{yellow}330 & \cellcolor{yellow}24 \\
    \textbf{YOLO} &  & 842 & 839 & 837 & 819 & 792 & 753 & 654 & 472 & 247 & 55 & 1 \\
    \textbf{Faster} &  & 1,324 & 1,309 & 1,280 & 1,241 & 1,149 & 994 & 718 & 462 & 204 & 58 & 3 \\
    \textbf{Retina} &  & 704 & 653 & 641 & 615 & 560 & 469 & 339 & 220 & 113 & 32 & 0 \\
    \midrule
    \textbf{DETR} & \multirow{4}{*}{\textbf{FP}} & \cellcolor{yellow}86 & \cellcolor{yellow}88 & \cellcolor{yellow}91 & \cellcolor{yellow}121 & \cellcolor{yellow}175 & \cellcolor{yellow}224 & \cellcolor{yellow}305 & \cellcolor{yellow}478 & \cellcolor{yellow}773 & \cellcolor{yellow}1,189 & \cellcolor{yellow}1,495 \\
    \textbf{YOLO} &  & 221 & 224 & 226 & 244 & 271 & 310 & 409 & 591 & 816 & 1,008 & 1,062 \\
    \textbf{Faster} &  & 230 & 245 & 274 & 313 & 405 & 560 & 836 & 1,092 & 1,350 & 1,496 & 1,551 \\
    \textbf{Retina} &  & 532 & 583 & 595 & 621 & 676 & 767 & 897 & 1,016 & 1,123 & 1,204 & 1,236 \\
    \midrule
    \textbf{DETR} & \multirow{4}{*}{\textbf{FN}} & \cellcolor{yellow}57 & \cellcolor{yellow}59 & \cellcolor{yellow}62 & \cellcolor{yellow}92 & \cellcolor{yellow}146 & \cellcolor{yellow}195 & \cellcolor{yellow}276 & \cellcolor{yellow}449 & \cellcolor{yellow}744 & \cellcolor{yellow}1,160 & \cellcolor{yellow}1,466 \\
    \textbf{YOLO} &  & 648 & 651 & 653 & 671 & 698 & 737 & 836 & 1,018 & 1,243 & 1,435 & 1,489 \\
    \textbf{Faster} &  & 166 & 181 & 210 & 249 & 341 & 496 & 772 & 1,028 & 1,286 & 1,432 & 1,487 \\
    \textbf{Retina} &  & 786 & 837 & 849 & 875 & 930 & 1,021 & 1,151 & 1,270 & 1,377 & 1,458 & 1,490 \\
    \midrule
    \textbf{DETR} & \multirow{4}{*}{\textbf{F1}} & \cellcolor{yellow}0.95 & \cellcolor{yellow}0.95 & \cellcolor{yellow}0.95 & \cellcolor{yellow}0.93 & \cellcolor{yellow}0.89 & \cellcolor{yellow}0.86 & \cellcolor{yellow}0.81 & \cellcolor{yellow}0.69 & \cellcolor{yellow}0.50 & \cellcolor{yellow}0.22 & \cellcolor{yellow}0.02 \\
    \textbf{YOLO} &  & 0.66 & 0.66 & 0.66 & 0.64 & 0.62 & 0.59 & 0.51 & 0.37 & 0.19 & 0.04 & 0 \\
    \textbf{Faster} &  & 0.87 & 0.86 & 0.84 & 0.82 & 0.75 & 0.65 & 0.47 & 0.30 & 0.13 & 0.04 & 0 \\
    \textbf{Retina} &  & 0.52 & 0.50 & 0.47 & 0.45 & 0.41 & 0.34 & 0.25 & 0.16 & 0.08 & 0.02 & 0 \\
    \midrule
    \multirow{4}{*}{\shortstack{\textbf{DETR} \\ \textbf{(signs)}}} & \textbf{TP} & 542 & 542 & 542 & 541 & 541 & 537 & 511 & 416 & 225 & 51 & 0 \\
    & \textbf{FP} & 5 & 5 & 5 & 6 & 6 & 10 & 36 & 131 & 322 & 496 & 547 \\
    & \textbf{FN} & 10 & 10 & 10 & 11 & 11 & 15 & 41 & 136 & 327 & 501 & 552 \\
    & \textbf{F1} & 0.99 & 0.99 & 0.99 & 0.98 & 0.98 & 0.98 & 0.93 & 0.76 & 0.41 & 0.09 & 0.00 \\
    \bottomrule
    \end{tabularx}
    \vspace{0.5em}
    \caption{Detection metrics across IoU thresholds. Utility pole results are shown for all models. Warning sign detection was only performed by DETR.}
    \label{tab:Test-Comparison}
    \end{table}

\subsection{Segmentation and Lean Angle Estimation}
The segmentation-enhanced DETR model was fine-tuned for 10 epochs using fixed detection weights. Segmentation masks enabled per-instance lean-angle estimation using OpenCV’s \texttt{fitLine()}, providing high angular accuracy. Of the 1,433 true-positive poles, lean angles were successfully estimated for 1,367 instances. The mean absolute error (MAE) was \SI{1.01}{\degree}, with a standard deviation of \SI{1.95}{\degree}. As summarised in Table~\ref{TAB: Combined Lean Angle Tables}, 98\% of predictions were within \SI{5}{\degree}, with 70.5\% within \SI{1}{\degree} and 90\% within \SI{2}{\degree}. Despite a maximum deviation of \SI{37.69}{\degree} for one outlier, estimation performance remained consistent across the test set.

\begin{table}[t]
\centering
\begin{minipage}[t]{0.48\textwidth}
\centering
\vspace{1.25ex}
\textbf{(a) Full Distribution}\\[0.5ex]
\begin{tabular}{|c|c|}
\hline
\textbf{Angle Error (°)} & \textbf{Count} \\
\hline
No Prediction & 66 \\
$0 \leq \theta \leq 5$ & 1,333 \\
$5 < \theta \leq 10$ & 26 \\
$10 < \theta \leq 15$ & 3 \\
$15 < \theta \leq 20$ & 3 \\
$25 < \theta \leq 30$ & 1 \\
$30 < \theta \leq 40$ & 1 \\
\hline
\end{tabular}
\end{minipage}
\hfill
\begin{minipage}[t]{0.48\textwidth}
\centering
\vspace{1.25ex}
\textbf{(b) Top Accuracy Breakdown}\\[0.5ex]
\begin{tabular}{|c|c|}
\hline
\textbf{Angle Error (°)} & \textbf{Count} \\
\hline
$0 \leq \theta \leq 1$ & 1,002 \\
$1 < \theta \leq 2$ & 203 \\
$2 < \theta \leq 3$ & 68 \\
$3 < \theta \leq 4$ & 35 \\
$4 < \theta \leq 5$ & 25 \\
\hline
\end{tabular}
\end{minipage}
\vspace{0.75em}
\caption{Lean angle error statistics across all poles (left) and top-performing subset (right).}
\label{TAB: Combined Lean Angle Tables}
\end{table}

On a filtered subset of 642 poles with similar height distribution to previous studies (mean height \SI{346.9}{px}), the model achieved \SI{92.8}{\percent} accuracy within \SI{1}{\degree}. This surpasses Alam et al.~\cite{Alam2020} by 22.3 percentage points, Zhu et al.~\cite{Zhu2019} by 11.85 points, and Kim et al.~\cite{Kim2021} by 17.8. Moreover, the filtered set was 3.7--11.5 times larger than those used in previous works, supporting the robustness and scalability of the proposed method.

\begin{figure}[t]
\centering
\includegraphics[width=0.4\linewidth]{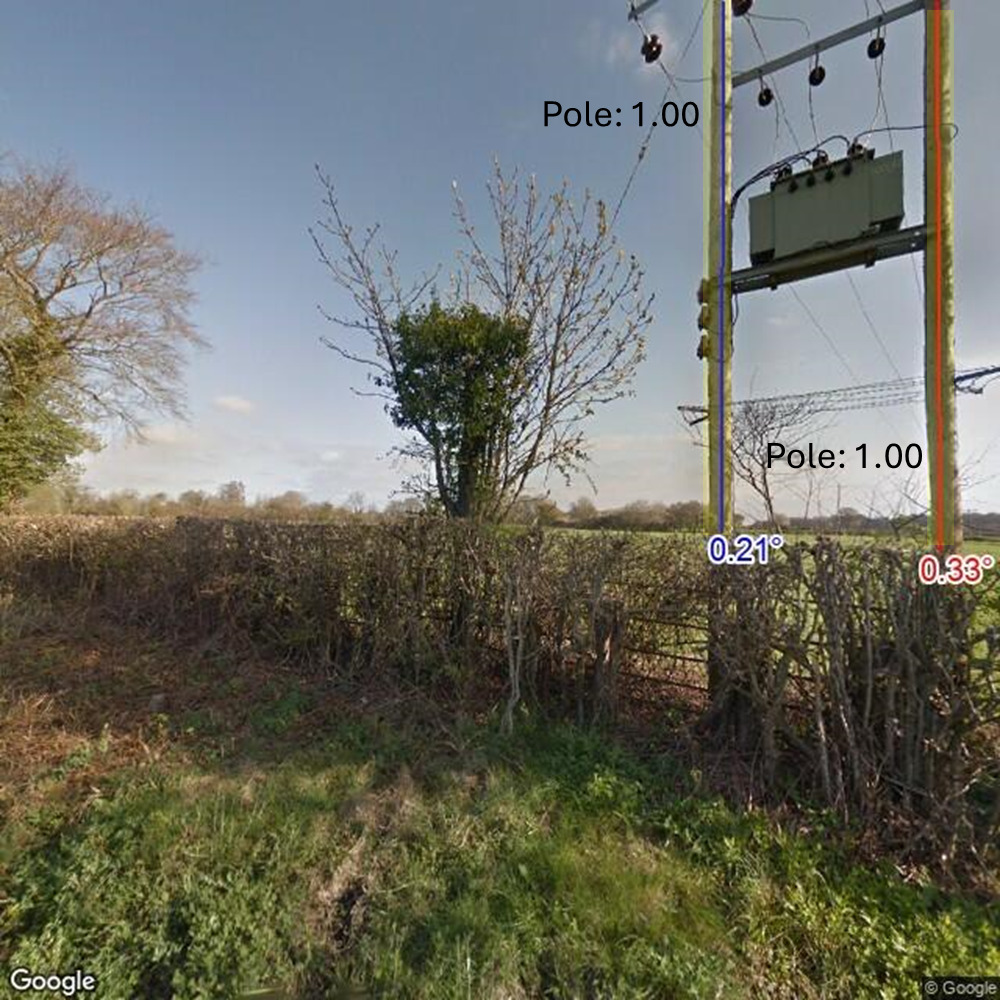}
\vspace{0.6em}
\caption{Examples of segmented utility poles with predicted confidence scores and estimated lean angles. The image highlights the extracted mask and corresponding orientation computed.}
\label{fig:segmentation_examples}
\end{figure}

Examples of the segmentation outputs, along with model confidence and estimated lean angles, are shown in Figure~\ref{fig:segmentation_examples}.

\section{Comparison with State-of-the-art models}
\label{sec:cr-addendum}

\noindent
To strengthen our baselines and test the \emph{generalisability} of the mask-based lean-angle estimator, we trained \yoloviii{}\cite{yolov8} and \dinodetr{}\cite{dinodetr} under the same five-fold protocol, data splits, training budget, and evaluation pipeline as in our original experiments. We additionally trained a \yoloviii{} segmentation model and applied the same mask-to-angle procedure (OpenCV \texttt{cv2.fitLine()}), enabling a like-for-like comparison against \detr{} with masks enabled.

\subsection{Cross-validation baselines (poles)}
Table~\ref{tab:cr_model_comparison_poles} reports pole detection performance (mean AP@0.5 across five folds with the mean peak epoch). Modern detectors are closely matched on poles.

\begin{table}[t]
\centering
\footnotesize
\begin{tabular}{lcc}
\toprule
\textbf{Model} & \textbf{Pole mAP} & \textbf{Peak Epoch} \\
\midrule
DETR       & 90.0\% & 91 \\
YOLOv8     & 89.7\% & 81 \\
DINO-DETR  & 89.3\% & 48 \\
\bottomrule
\end{tabular}
\caption{Cross-validation performance on wooden utility pole detection (mean AP@0.5 over five folds).}
\label{tab:cr_model_comparison_poles}
\end{table}

Across five folds the three models are essentially tied on poles (AP@0.5: \detr{} 90.0\%, \yoloviii{} 89.7\%, \dinodetr{} 89.3\%), closing the gap to our earlier baselines and confirming that strong localisation on \emph{OHL-UK} is not an artefact of weak comparators. \dinodetr{} peaks earlier (48 epochs) than \yoloviii{} (81) and \detr{} (91), affecting wall-clock time rather than ranking.

\subsection{Test-set analysis across IoU thresholds}
Using each method’s best confidence by AUC(F1), Tables~\ref{tab:cr_iou_sweep_additional_poles} and \ref{tab:cr_iou_sweep_additional_signs} compacts the held-out test-set IoU sweeps to F1, and also lists TP/FP/FN at IoU\,{=}\,0.5. On \emph{poles}, \yoloviii{} attains the highest F1 at IoU\,{=}\,0.5 (0.93), with \detr{} competitive (0.86). On \emph{signs}, \detr{} and \yoloviii{} are tied at IoU\,{=}\,0.5 (0.98), with \yoloviii{} tighter at higher IoUs. Counts at IoU$=0.5$ corroborate this: on \emph{poles} YOLOv8 has more TP and fewer errors than \detr{} ($+44$ TP, $-119$ FP, $-109$ FN vs.\ \detr{}), while DINO-DETR trails (1030/401/395); for \emph{signs} the counts are near-identical (DETR 537/10/15 vs.\ YOLOv8 532/12/15).

\begin{table}[H]
\centering
\footnotesize
\setlength{\tabcolsep}{3pt}
\renewcommand{\arraystretch}{0.95}
\begin{tabularx}{\textwidth}{l|*{11}{C}}
\toprule
\textbf{Model} & \textbf{0.0} & \textbf{0.1} & \textbf{0.2} & \textbf{0.3} & \textbf{0.4} & \textbf{0.5} & \textbf{0.6} & \textbf{0.7} & \textbf{0.8} & \textbf{0.9} & \textbf{1.0} \\
\midrule
DETR   & 0.95 & 0.95 & 0.95 & 0.93 & 0.89 & 0.86 & 0.81 & 0.69 & 0.50 & 0.22 & 0.02 \\
YOLOv8 & 0.95 & 0.95 & 0.95 & 0.94 & 0.94 & 0.93 & 0.89 & 0.81 & 0.56 & 0.17 & 0.00 \\
DINO   & 0.88 & 0.74 & 0.73 & 0.73 & 0.73 & 0.72 & 0.70 & 0.64 & 0.50 & 0.18 & 0.00 \\
\bottomrule
\end{tabularx}
\caption{Poles: F1 across IoU thresholds on the test set (best confidence by AUC(F1)).}
\label{tab:cr_iou_sweep_additional_poles}
\end{table}

\begin{table}[H]
\centering
\footnotesize
\setlength{\tabcolsep}{3pt}
\renewcommand{\arraystretch}{0.95}
\begin{tabularx}{\textwidth}{l|*{11}{C}}
\toprule
\textbf{Model} & \textbf{0.0} & \textbf{0.1} & \textbf{0.2} & \textbf{0.3} & \textbf{0.4} & \textbf{0.5} & \textbf{0.6} & \textbf{0.7} & \textbf{0.8} & \textbf{0.9} & \textbf{1.0} \\
\midrule
DETR   & 0.99 & 0.99 & 0.99 & 0.98 & 0.98 & 0.98 & 0.93 & 0.76 & 0.41 & 0.09 & 0.00 \\
YOLOv8 & 0.98 & 0.98 & 0.98 & 0.98 & 0.98 & 0.98 & 0.95 & 0.84 & 0.50 & 0.08 & 0.00 \\
DINO   & 0.85 & 0.49 & 0.49 & 0.49 & 0.49 & 0.48 & 0.47 & 0.41 & 0.23 & 0.04 & 0.00 \\
\bottomrule
\end{tabularx}
\caption{Signs: F1 across IoU thresholds on the test set (best confidence by AUC(F1)).}
\label{tab:cr_iou_sweep_additional_signs}
\end{table}

\vspace{3pt}
\begin{table}[H]
\centering
\footnotesize
\renewcommand{\arraystretch}{0.95}
\begin{tabular}{lcc}
\toprule
 & \textbf{Poles @ IoU=0.5 (TP/FP/FN)} & \textbf{Signs @ IoU=0.5 (TP/FP/FN)} \\
\midrule
DETR   & 1295 / 224 / 195 & 537 / 10 / 15 \\
YOLOv8 & 1339 / 105 / 86  & 532 / 12 / 15 \\
DINO   & 1030 / 401 / 395 & 261 / 277 / 286 \\
\bottomrule
\end{tabular}
\caption{TP/FP/FN counts at IoU=0.5 (best confidence by AUC(F1)).}
\label{tab:cr_counts_iou05}
\end{table}

\subsection{Segmentation and lean-angle estimation}
We trained a \yoloviii{} segmentation model and evaluated it with the same mask-to-angle pipeline as \detr{}+seg (\texttt{cv2.fitLine}). On the test set with IoU$\geq$0.5 matching, \yoloviii{}-seg produced 1{,}313/1{,}490 matches; MAE $=0.645^\circ$ (median $0.339^\circ$, P90 $1.453^\circ$), with $83.3\%$ (1{,}093/1{,}313) within $1^\circ$ and $99.0\%$ (1{,}300/1{,}313) within $5^\circ$ (13 $>5^\circ$). For \detr{}+seg, angles were produced for 1{,}367/1{,}433 detections (4.6\% no prediction), MAE $=1.01^\circ$, with $73.3\%$ (1{,}002/1{,}367) within $1^\circ$ and $97.5\%$ (1{,}333/1{,}367) within $5^\circ$ (34 $>5^\circ$). 
GT angles come from mask-derived \texttt{fitLine} estimates, so errors are mask-to-mask, not absolute tilt. \yoloviii{}-seg is sharper per instance; \detr{}+seg covers more. Use \detr{} as the backbone and prefer/fuse \yoloviii{}-seg angles when both masks exist.

\section{Discussion and Limitations}
\label{sec:discussion}
The DETR-based model demonstrates strong performance in detecting utility poles and warning signs within complex street-level imagery. However, several limitations should be considered when interpreting results or planning deployment.

\textbf{Viewpoint and Occlusion.} Detection performance degrades in scenarios involving occlusion (e.g.\ vegetation, vehicles) or non-frontal viewpoints, due to reliance on monocular, single-perspective imagery.

\textbf{Domain Generalisation.} The dataset comprises scenes from UK suburban and rural environments. Application to different geographic regions or infrastructure types may require domain adaptation or additional training data.

\textbf{Resolution Sensitivity.} Detection accuracy diminishes for small or low-resolution signs, particularly those under \SI{20}{px} in height. Enhancing performance may require multi-scale feature aggregation or super-resolution modules, at increased computational cost.

\textbf{Segmentation Dependency.} Lean-angle estimation is sensitive to segmentation quality. Inaccurate masks introduce angular error. Improvements may be achieved through uncertainty modelling, confidence-weighted filtering, or incorporating depth cues.

\textbf{Annotation Noise.} Despite quality control, annotation errors persist—especially in small bounding boxes and fine-grained angle labels. These introduce noise into both detection and regression objectives.

Given these constraints, deployment in safety-critical inspection workflows should incorporate human oversight or redundancy to ensure robustness and interpretability.

\section{Conclusion}
\label{sec:conclusion}
We have presented a unified deep learning framework for detecting and structurally assessing wooden utility poles and electrical warning signs from ground-level imagery. Our system, based on a DETR backbone with a segmentation head, achieves state-of-the-art results across multiple metrics, including mean average precision and lean angle estimation accuracy. It significantly outperforms popular object detectors such as RetinaNet, YOLOv3-Tiny, and Faster R-CNN.

The contributions of this work include: (1) the OHL-UK dataset, an open curated collection of over 6,000 poles and 1,800 signs; (2) a DETR-based detection model tailored for long, thin objects; and (3) a lean-angle estimation pipeline achieving sub-degree accuracy from monocular images. These innovations support scalable, low-cost infrastructure inspection and could improve safety, consistency, and coverage in utility pole monitoring.

Future directions include extending the model to detect additional structural features (e.g.\ rot, cracks, equipment types), incorporating 3D reasoning, and validating generalisation across geographies. The code, annotations, and trained models will be released upon publication to facilitate further research and responsible deployment.

\section*{Acknowledgements}
This work was supported by UK Power Networks (UKPN), who funded the project and provided the geographic coordinates of their utility poles, enabling the collection of ground-level imagery used in this study.

\bibliography{references}

\begin{thebibliography}{18}
\providecommand{\natexlab}[1]{#1}
\providecommand{\url}[1]{\texttt{#1}}
\expandafter\ifx\csname urlstyle\endcsname\relax
  \providecommand{\doi}[1]{doi: #1}\else
  \providecommand{\doi}{doi: \begingroup \urlstyle{rm}\Url}\fi

\bibitem[Arc()]{Archie_West}
Introduction -- {ARCHIE-WeSt} documentation.
\newblock \url{https://docs.hpc.strath.ac.uk/user-guide/}.
\newblock Accessed 2025.

\bibitem[Alam et~al.(2020)Alam, Zhu, Tokgoz, Zhang, and Hwang]{Alam2020}
Md~Morshedul Alam, Zanbo Zhu, Berna~Eren Tokgoz, Jing Zhang, and Seokyon Hwang.
\newblock Automatic assessment and prediction of the resilience of utility
  poles using unmanned aerial vehicles and computer vision techniques.
\newblock \emph{International Journal of Disaster Risk Science}, 11\penalty0
  (1):\penalty0 119--132, 2020.
\newblock \doi{10.1007/s13753-020-00254-1}.
\newblock URL \url{https://doi.org/10.1007/s13753-020-00254-1}.

\bibitem[Carion et~al.(2020)Carion, Massa, Synnaeve, Usunier, Kirillov, and
  Zagoruyko]{Carion2020}
Nicolas Carion, Francisco Massa, Gabriel Synnaeve, Nicolas Usunier, Alexander
  Kirillov, and Sergey Zagoruyko.
\newblock End-to-end object detection with transformers.
\newblock In Andrea Vedaldi, Horst Bischof, Thomas Brox, and Jan-Michael Frahm,
  editors, \emph{Computer Vision -- ECCV 2020}, volume 12346 of \emph{Lecture
  Notes in Computer Science}, pages 213--229. Springer, 2020.
\newblock \doi{10.1007/978-3-030-58452-8_13}.
\newblock URL
  \url{https://link.springer.com/chapter/10.1007/978-3-030-58452-8_13}.

\bibitem[Cubuk et~al.(2019)Cubuk, Zoph, Mane, Vasudevan, and Le]{Cubuk2019}
Ekin~D. Cubuk, Barret Zoph, Dandelion Mane, Vijay Vasudevan, and Quoc~V. Le.
\newblock Autoaugment: Learning augmentation strategies from data.
\newblock In \emph{Proceedings of the IEEE/CVF Conference on Computer Vision
  and Pattern Recognition}, pages 113--123, 2019.
\newblock \doi{10.1109/CVPR.2019.00020}.

\bibitem[Dickinson(2025)]{ohluk2025}
Carl Dickinson.
\newblock {OHL-UK}: Wooden utility pole and electrical sign corpus with trained
  detection and segmentation models, 2025.
\newblock URL
  \url{https://doi.org/10.15129/df7cc895-1091-4cd6-bf2c-7b5ebd203e55}.
\newblock Dataset openly available under CC BY 4.0 licence.

\bibitem[Dutta and Zisserman(2019)]{Dutta2019}
Abhishek Dutta and Andrew Zisserman.
\newblock The {VGG} image annotator ({VIA}).
\newblock \emph{arXiv preprint arXiv:1904.10699}, 2019.
\newblock URL \url{https://arxiv.org/abs/1904.10699}.

\bibitem[Gon{\c c}alves and Carvalho(2013)]{Goncalves2013}
Rog{\'e}rio Gon{\c c}alves and Jo{\~a}o Carvalho.
\newblock Review and latest trends in mobile robots used on power transmission
  lines.
\newblock \emph{International Journal of Advanced Robotic Systems},
  10:\penalty0 1--14, 2013.
\newblock \doi{10.5772/56791}.
\newblock URL \url{https://doi.org/10.5772/56791}.

\bibitem[Kim et~al.(2021)Kim, Kamari, Lee, and Ham]{Kim2021}
Jaeyoon Kim, Mirsalar Kamari, Seulbi Lee, and Youngjib Ham.
\newblock Large-scale visual data--driven probabilistic risk assessment of
  utility poles regarding the vulnerability of power distribution
  infrastructure systems.
\newblock \emph{Journal of Construction Engineering and Management},
  147\penalty0 (10):\penalty0 04021121, 2021.
\newblock \doi{10.1061/(ASCE)CO.1943-7862.0002153}.

\bibitem[Liu et~al.(2019)Liu, Zhang, Zhao, Wiliem, Astin-Walmsley, and
  Lovell]{Liu2019}
Liangchen Liu, Teng Zhang, Kun Zhao, Arnold Wiliem, Kieren Astin-Walmsley, and
  Brian Lovell.
\newblock Deep inspection: An electrical distribution pole parts study {VIA}
  deep neural networks.
\newblock In \emph{2019 IEEE International Conference on Image Processing
  (ICIP)}, pages 4170--4174. IEEE, 2019.
\newblock \doi{10.1109/ICIP.2019.8803415}.
\newblock URL \url{https://ieeexplore.ieee.org/document/8803415}.

\bibitem[Mounsaveng et~al.(2021)Mounsaveng, Laradji, Ben~Ayed, Vazquez, and
  Pedersoli]{Mounsaveng2021}
Saypraseuth Mounsaveng, Issam Laradji, Ismail Ben~Ayed, David Vazquez, and
  Marco Pedersoli.
\newblock Learning data augmentation with online bilevel optimization for image
  classification.
\newblock In \emph{2021 IEEE Winter Conference on Applications of Computer
  Vision (WACV)}, pages 1690--1699, 2021.
\newblock \doi{10.1109/WACV48630.2021.00173}.

\bibitem[Sharma et~al.(2015)Sharma, Adithya, Dutta, and
  Balamuralidhar]{Sharma2015}
Hrishikesh Sharma, V.~Adithya, Tanima Dutta, and P.~Balamuralidhar.
\newblock Image analysis-based automatic utility pole detection for remote
  surveillance.
\newblock In \emph{2015 International Conference on Digital Image Computing:
  Techniques and Applications (DICTA)}, pages 1--7. IEEE, 2015.
\newblock \doi{10.1109/DICTA.2015.7371267}.
\newblock URL \url{https://ieeexplore.ieee.org/document/7371267}.

\bibitem[{UK Power Networks}(2017)]{UKPN2017-FutureSmart}
{UK Power Networks}.
\newblock Future smart: Consultation report.
\newblock Technical report, UK Power Networks, 2017.
\newblock URL
  \url{https://media.umbraco.io/uk-power-networks/2etfstsb/futuresmart_consultationreport_.pdf}.

\bibitem[Wen()]{GSVAPI}
Richard Wen.
\newblock google\_streetview: Google street view image api command-line tool
  and python module (v1.2.3).
\newblock \url{https://rrwen.github.io/google_streetview/}.
\newblock Accessed 2025.

\bibitem[Whitworth et~al.(2001)Whitworth, Duller, Jones, and Earp]{Jones2001}
C.~C. Whitworth, A.~W.~G. Duller, D.~I. Jones, and G.~K. Earp.
\newblock Aerial video inspection of overhead power lines.
\newblock \emph{Power Engineering Journal}, 15\penalty0 (1):\penalty0 25--32,
  2001.
\newblock \doi{10.1049/pe:20010103}.
\newblock URL
  \url{https://digital-library.theiet.org/doi/10.1049/pe%3A20010103}.

\bibitem[Yaseen(2024)]{yolov8}
Muhammad Yaseen.
\newblock What is {YOLOv8}: An in-depth exploration of the internal features of
  the next-generation object detector, 2024.
\newblock URL \url{https://arxiv.org/abs/2408.15857}.

\bibitem[Zhang et~al.(2022)Zhang, Li, Liu, Zhang, Su, Zhu, Ni, and
  Shum]{dinodetr}
Hao Zhang, Feng Li, Shilong Liu, Lei Zhang, Hang Su, Jun Zhu, Lionel~M. Ni, and
  Heung-Yeung Shum.
\newblock {DINO}: {DETR} with improved denoising anchor boxes for end-to-end
  object detection, 2022.
\newblock URL \url{https://arxiv.org/abs/2203.03605}.

\bibitem[Zhang et~al.(2018)Zhang, Witharana, Li, Zhang, Li, and
  Parent]{UsingDL}
Weixing Zhang, Chandi Witharana, Weidong Li, Chuanrong Zhang, Xiaojiang Li, and
  Jason Parent.
\newblock Using deep learning to identify utility poles with crossarms and
  estimate their locations from google street view images.
\newblock \emph{Sensors}, 18\penalty0 (8):\penalty0 2484, 2018.
\newblock \doi{10.3390/s18082484}.
\newblock URL \url{https://www.mdpi.com/1424-8220/18/8/2484}.

\bibitem[Zhu et~al.(2019)Zhu, Zhang, Alam, Tokgoz, and Hwang]{Zhu2019}
Zanbo Zhu, Jing Zhang, Md~Morshedul Alam, Berna~Eren Tokgoz, and Seokyon Hwang.
\newblock Automatic utility pole inclination angle measurement using unmanned
  aerial vehicle and deep learning.
\newblock In \emph{IISE Annual Conference and Expo 2019}, 2019.

\end{thebibliography}
\end{document}